\documentclass{article}
\usepackage{fancyhdr}

\usepackage{report}

\usepackage{soul}
\setcitestyle{numbers,square,sort&compress}
\usepackage[utf8]{inputenc} 
\usepackage[T1]{fontenc}    
\usepackage{hyperref}       
\usepackage{url}            
\usepackage{booktabs}       
\usepackage{amsfonts}       
\usepackage{fancyhdr}       

\usepackage{nicefrac}       
\usepackage{microtype}      
\usepackage{graphicx}
\usepackage{pifont}
\usepackage{multirow}
\usepackage{CJKutf8}
\definecolor{darkmagenta}{rgb}{0.56, 0.0, 1.0}
\definecolor{softyellow}{rgb}{1.0, 0.92, 0.3} 
\definecolor{LightAquamarine}{rgb}{0.75, 1.0, 0.8} 
\definecolor{FireBrick}{RGB}{178,34,34}
\definecolor{MediumPurple}{RGB}{147,112,219}

\definecolor{uclablue}{rgb}{0.15, 0.45, 0.68}
\hypersetup{
    breaklinks,
    colorlinks=true,
    citecolor={darkmagenta},
    linkcolor={uclablue},
    urlcolor={uclablue}
}
\usepackage{wrapfig}
\usepackage{float}
\usepackage{subcaption}
\usepackage{placeins}
\usepackage{lipsum} 
\usepackage{tcolorbox}
\usepackage{amsmath}
\usepackage{amssymb}
\usepackage{utfsym}
\usepackage{fontawesome}
\usepackage{xspace}
\usepackage{enumitem}
\usepackage{multirow} 
\tcbuselibrary{breakable}
\usepackage{enumitem}
\usepackage{colortbl}
\usepackage{fancyhdr}

\tcbuselibrary{skins}
\usepackage{transparent}

\definecolor{njuPurple}{RGB}{220,205,230}     
\definecolor{njuPurpleLight}{RGB}{250,245,252}   

\newtcolorbox{abstractbox}{
    colback=njuPurpleLight,   
    colframe=njuPurple,       
    boxrule=1pt,              
    arc=4mm,                  
    left=8pt,                 
    right=8pt,                
    top=8pt,                  
    bottom=8pt,               
    opacityback=0.95
}

\definecolor{PromptNavy}{HTML}{355C7D}
\definecolor{PromptPurple}{HTML}{8064A2}
\definecolor{PromptBlueBG}{HTML}{F4F8FB}
\definecolor{PromptPurpleBG}{HTML}{F8F5FB}
\newtcolorbox{promptbox}[1][]{
  enhanced,
  breakable,
  colback=PromptPurpleBG,
  colframe=PromptPurple,
  coltitle=black,
  fonttitle=\bfseries\small,
  title={#1},
  boxrule=0.6pt,
  arc=2pt,
  left=7pt,
  right=7pt,
  top=6pt,
  bottom=6pt,
  before skip=6pt,
  after skip=8pt,
  fontupper=\footnotesize
}

\usepackage{xspace}
\usepackage{makecell}
\newcommand{\name}{CodeCap-Bench\xspace}

\title{\name: Can Code Serve as a Good Video Caption?}

\author{%
    Jiafu Tang$^{1,2,*,\spadesuit}$,
    Yifan Xu$^{1,2,*}$,
    Eric Tang$^{3}$,
    Junjie Zhao$^{4}$,
    Shipeng Luo$^{5}$ \\
    Te Sun$^{4}$,
    Haoxuan Hu$^{1}$,
    Yiyan Ji$^{1}$,
    Xinlong Chen$^{5}$,
    Zihao Wang$^{2}$ \\
    Yang Shi$^{6,2,\dagger}$,
    Yuanxing Zhang$^{2,\spadesuit}$,
    Pengfei Wan$^{2}$,
    Jiaheng Liu$^{1,\dagger}$ \\[0.6em]
    $^{1}$NJU \quad
    $^{2}$Kling AI \quad
    $^{3}$TJU \quad
    $^{4}$SJTU \quad
    $^{5}$CASIA \quad
    $^{6}$PKU \\[0.4em]
    Correspondence: \texttt{liujiaheng@nju.edu.cn} \\[0.4em]
    Project page: \url{https://codecap-bench.github.io/}
}

\date{October 2026}

\begin{document}

\maketitle
\begingroup
\renewcommand{\thefootnote}{}
\footnotetext{%
    $^{*}$Equal Contribution.\quad
    $^{\dagger}$Corresponding Author.\quad
    $^{\spadesuit}$Project Lead.
}
\endgroup

\begin{abstractbox}
\begin{center}
\textbf{\Large Abstract}
\end{center}

Video captioning aims to preserve visual information through compact descriptions, predominantly using natural language.
However, precisely expressing fine-grained spatial structures and temporal dynamics in natural language often leads to lengthy and potentially ambiguous descriptions.
In this work, we explore \textbf{Code as Video Caption}, a paradigm where executable programs serve as a complementary representation of visual content through graphical primitives, layout rules, and functions of time.
To examine whether code captions effectively preserve video information, we compare them with natural language captions on downstream question answering and video reconstruction tasks.
Our results show that code captions achieve superior reasoning accuracy and reconstruction fidelity.
Despite these advantages, realizing the potential of code captioning depends on whether current multimodal large language models (MLLMs) can generate faithful code captions.
To systematically assess this capability, we introduce \textbf{CodeCap-Bench}, which comprises $323$ videos spanning five categories of structured 2D content, along with fine-grained annotations and a multi-level evaluation framework.
Extensive experiments reveal that while leading proprietary MLLMs demonstrate promising code captioning capabilities, they still struggle to faithfully capture complex temporal dynamics, with open-source models lagging substantially behind.

\end{abstractbox}

\section{Introduction}
\label{sec:intro}

Video captioning~\citep{chen2026avocado, chen2026diadem, chen2026pigeons, chen2026capricorn, li2026if, liu2026refcaptioner, shi2025mavors, wang2026omnicap, wang2026avscaporchestratingaudiovisualsynergy, Wu_2025_CVPR, xu2026percepcap} aims to preserve visual information in videos through compact descriptions.
Existing methods mainly rely on natural language to describe objects, attributes, actions, and events~\citep{chu2025fine, ma2026omni, xue2025progress}.
However, videos also contain fine-grained spatial structures and temporal dynamics, including complex layouts, motion trajectories, and coordinated changes among multiple objects.
Precisely expressing this information in natural language often requires many positional, temporal, and relational constraints, making descriptions lengthy and potentially ambiguous.
This motivates us to explore \textbf{Code as Video Caption}, where executable programs provide a complementary way to represent visual content through graphical primitives, layout rules, and temporal logic~\citep{chen2025code2video, rodriguez2025starvector, zhang2026code2worlds}.

\begin{figure*}[t]
    \centering
    \includegraphics[width=\textwidth]{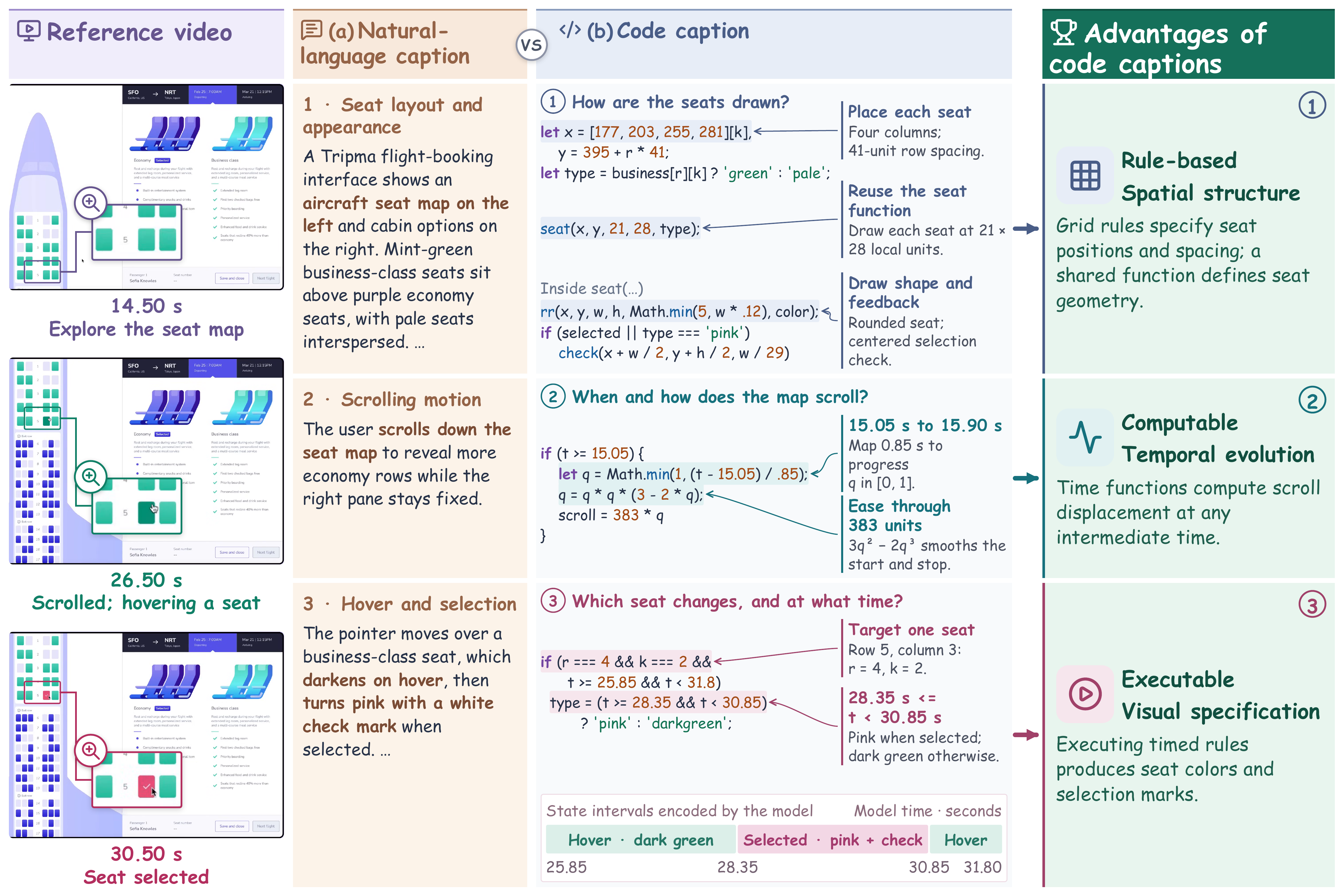}
    \caption{\textbf{Illustration of video captioning with executable code.} The reference video shows a flight seat-selection interface. (a) and (b) present natural language and code captions of the same segment, respectively. The code caption encodes spatial structure, temporal dynamics, and visual state changes through executable rules and functions.}
    \label{fig:code_as_video_caption}
\end{figure*}

As illustrated in Figure~\ref{fig:code_as_video_caption}, executable code has three complementary advantages compared with natural language descriptions in representing video information. First, code captures rule-based spatial structure. In the flight seat-selection interface, grid rules specify seat positions and spacing, while a shared drawing function defines seat geometry, compactly representing repeated elements and their spatial relationships. Second, code supports computable temporal evolution. While the natural language caption states that “the seat map scrolls downward,” time functions specify the scrolling interval, displacement, and easing, allowing the scroll position to be computed at any intermediate time. Finally, code provides an executable visual specification. Conditional logic specifies which seat changes color and when a selection mark appears, and executing these rules produces the corresponding visual states. Together, these properties make spatial layouts, motion, and state changes explicit, reduce ambiguity in how visual details are interpreted, and allow the rendered video to be directly compared with the reference video.

To examine the practical value of code as video caption, we compare code and natural language captions through question answering and video reconstruction (Section~\ref{subsubsec:code_vs_nl}). Under the evaluated settings, videos rendered from code captions support more accurate question answering and better preserve the appearance, motion patterns, and spatial structure of the original videos in reconstruction, providing evidence of the potential of code captions to preserve and utilize video information.

Despite this promise, a dedicated benchmark for this task is still lacking, and the ability of current multimodal large language models~\citep{anthropic2026opus5,chen2026videochat,openai2026gpt6astra,qwen2026qwen38max,wang2026beacon,wang2026monet,zai2026glm53flash} to generate faithful code captions has yet to be systematically evaluated. In particular, how accurately models can represent spatial layouts, motion trajectories, and state changes, and how these capabilities vary across content types and generation settings, remain open questions.

To address these questions, we introduce \textbf{\name}, a benchmark for systematically evaluating the code captioning capabilities of MLLMs.
We focus on structured 2D videos whose visual content can be represented programmatically.
\name comprises $323$ videos across five categories: desktop and application interfaces, 2D games, visual design and media animation, data and scientific visualization, and icons and abstract motion graphics.
We provide fine-grained focal-point annotations of key static and dynamic content, together with object-level spatio-temporal annotations.
By rendering generated code into videos, we evaluate caption fidelity along three complementary dimensions: \textit{visual similarity}, \textit{object-level spatio-temporal IoU}, and \textit{fine-grained visual information retention}.
This framework enables systematic assessment of how faithfully models preserve visual appearance, spatial structure, and temporal dynamics through executable code.

We conduct extensive experiments to investigate the ability of current MLLMs to generate code captions.
Our evaluation reveals substantial differences across models, with dynamic visual information consistently less faithfully preserved than static content.
We further examine the effects of agentic workflows, thinking effort, and generation cost on caption fidelity and reliability.
These results reveal the capabilities and limitations of current models in generating faithful code captions and provide empirical evidence to help connect video understanding with programmatic generation.

Our contributions can be summarized as follows:
\begin{itemize}
    \item We explore \textbf{Code as Video Caption}, investigating executable code as a complementary representation for preserving fine-grained visual information, spatial structure, and temporal dynamics. Code captions have an explicit visual interpretation that can be checked through rendering, with question-answering and video reconstruction comparisons providing further empirical support for their value.
    \item We introduce \textbf{\name}, a benchmark comprising $323$ videos across five categories of structured 2D content, with fine-grained focal-point and object-level spatio-temporal annotations. We establish a multi-level evaluation framework covering global visual similarity, object-level spatio-temporal alignment, and fine-grained visual information retention.
    \item We systematically evaluate MLLMs across generation settings, finding that leading proprietary models show promise but struggle with complex temporal dynamics, while open-source models lag substantially behind. We also analyze the relationships among caption fidelity, generation reliability, and cost.
\end{itemize}

\section{Related Work}
\label{sec:related_work}

\paragraph{Fine-grained video captioning and evaluation.}
Recent video captioning studies increasingly focus on fine-grained visual information beyond high-level semantic summaries.
DREAM-1K~\citep{wang2024tarsier} provides detailed descriptions of events, actions, and motion, VDC~\citep{chai2025auroracap} structures captions around objects, backgrounds, and camera motion, and CaReBench~\citep{xu2026carebench} distinguishes spatial and temporal information for captioning and retrieval.
Caption evaluation has also shifted toward assessing preserved visual information rather than text similarity alone.
VDCscore~\citep{chai2025auroracap} and VidCapBench~\citep{chen2025vidcapbench} use reference-caption-derived question-answer pairs and targeted video-attribute questions, respectively, while CapQuiz~\citep{wang2026putting} adopts reference-free multiple-choice questions.
Whereas these methods evaluate natural language captions, \name evaluates executable code captions through rendered videos, measuring overall visual similarity, object-level spatio-temporal IoU, and fine-grained visual information retention.

\paragraph{Vision2Code.}
Recent studies explore reconstructing visual content through executable code.
For static images, Design2Code~\citep{si2025design2code}, ChartMimic~\citep{yang2025chartmimic}, and Image2Struct~\citep{roberts2024image2struct} evaluate code generation from webpage screenshots, scientific charts, and images of structured content, respectively.
Other studies examine how code preserves visual information.
VCode~\citep{lin2025vcode} evaluates semantic information retained in SVG representations through CodeVQA, while CodePercept~\citep{guan2026codepercept} improves STEM image perception through code-grounded training and assesses visual understanding through reconstruction code generation.
For videos, Animation2Code~\citep{ji2026animation2code} evaluates web animation reconstruction using HTML, CSS, and JavaScript through appearance and temporal metrics.
BVB~\citep{tang2026bvb} reconstructs indoor videos as animated Blender scenes, evaluated through spatio-temporal questions and representation similarity.
Related studies reconstruct interactive webpages from user interaction videos, assessing appearance, interaction states, and functionality~\citep{chen2026iwr,dai2026webvr,xu2026video2code}.
While these studies focus on specific domains, \name evaluates code-based video captioning across diverse structured 2D videos, emphasizing fine-grained visual information retention and spatio-temporal fidelity.

\section{CodeCap-Bench}
\label{sec:benchmark}

\subsection{Task Formulation}
We formulate video captioning with code as the task of expressing visual content and its temporal evolution through executable programs.
Given a reference video $V = \{I_t\}_{t \in [0, T]}$ of duration $T$, the goal is to generate a code caption $C$ that captures its visual elements, spatial layouts, and temporal dynamics.
Unlike natural language captions, which describe visual content through text, code captions explicitly specify how visual elements are constructed, arranged, and updated over time using graphical primitives, layout rules, and temporal logic.
We focus on the pure-code setting, where all visual content must be generated programmatically without reusing reference frames or external visual assets.
Executing the code through a rendering engine $\mathcal{R}$ produces a video $\hat{V} = \mathcal{R}(C)$, whose visual states should align with those of the reference video throughout the timeline.
Code caption fidelity is therefore evaluated by comparing $\hat{V}$ with $V$ in terms of visual appearance, spatial structure, and temporal dynamics.

\subsection{Evaluation Framework}
\label{subsec:evaluation_framework}

We evaluate code caption fidelity along three complementary dimensions: \textit{visual similarity}, \textit{object-level spatio-temporal IoU}, and \textit{fine-grained visual information retention}.
These metrics assess the consistency between the reference and rendered videos at different levels.
Visual similarity measures global similarity in visual content, appearance, and motion patterns, while spatio-temporal IoU evaluates the spatial and temporal alignment of key objects.
Fine-grained visual information retention further examines how well important static content and dynamic processes are preserved.

\begin{figure}[t]
    \centering
    \IfFileExists{figures/codecap_data_construction.pdf}{
        \includegraphics[width=\linewidth]{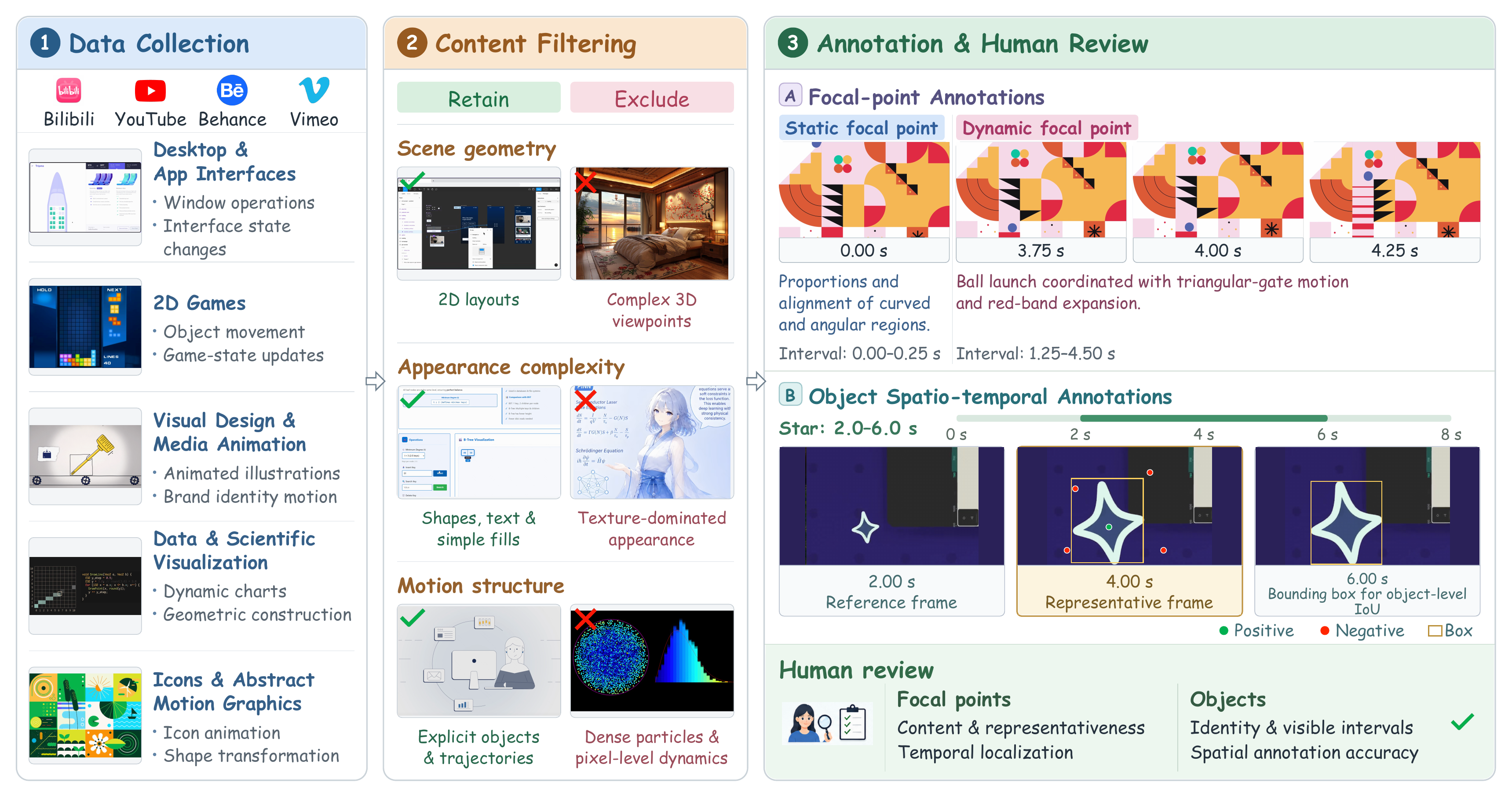}
    }{
        \fbox{\parbox[c][4cm][c]{0.9\linewidth}{\centering Data construction pipeline}}
    }
    \caption{\textbf{The data construction pipeline of CodeCap-Bench.} The pipeline comprises three stages: multi-source video collection, content filtering for structured 2D videos, and fine-grained annotation with human review, covering static and dynamic focal points and object-level spatio-temporal information.}
    \label{fig:codecap_data_construction}
\end{figure}

\paragraph{Visual similarity.}
Visual similarity measures the overall visual consistency between the reference and rendered videos.
We apply identical temporal sampling and preprocessing to both videos and extract their representations using the pretrained V-JEPA 2.1~\citep{murlabadia2026vjepa21} ViT-Gigantic-384 model.
We then compute the cosine similarity between the resulting video features to assess their global similarity in visual content, appearance, and motion patterns.

\paragraph{Object-level spatio-temporal IoU.}
Object-level spatio-temporal IoU measures the spatial and temporal alignment of key objects between the reference and rendered videos.
For each annotated reference object, GPT-6-sol~\citep{openai2026gpt6sol} identifies its visible interval in the rendered video, selects representative frames, and generates bounding boxes along with positive and negative point prompts to guide video segmentation with SAM 3.1~\citep{meta2026sam31}.
We extract per-frame bounding boxes from the segmentation results and compute their IoU with the reference annotations on a shared timeline.
The IoU is averaged over frames where the object is visible in either video, with a score of zero when the object is absent from the rendered video.
We report the object score by averaging object-level IoU scores first within each video, then across all videos.

\paragraph{Fine-grained visual information retention.}
Fine-grained visual information retention measures how faithfully a code caption preserves the static and dynamic visual content specified by annotated focal points.
For each focal point, GPT-6-sol~\citep{openai2026gpt6sol} receives the corresponding reference segment, the focal point's textual description, and the full rendered video.
Based on these inputs, it assigns an integer score from $0$ to $5$ to indicate how faithfully the specified content is reproduced.
Static focal points concern appearance, layout, and spatial relationships, while dynamic focal points concern state changes, motion processes, pacing, and temporal coordination.
Higher scores indicate more faithful information retention, with $0$ indicating that the relevant content is not substantively reproduced.
As with object-level spatio-temporal IoU, scores are averaged within each video and then across all videos.

\subsection{Data Construction}
\label{subsec:data_construction}

We construct \name through a three-stage pipeline comprising multi-source video collection, content filtering, and fine-grained annotation with human review, as illustrated in Figure~\ref{fig:codecap_data_construction}.
We focus on structured 2D videos, where fine-grained visual details, spatial layouts, and temporal dynamics are central to the visual content.
The explicit structure of these videos facilitates fine-grained annotation and comparison between reference and rendered videos.
These videos exhibit diverse and complex visual structures and dynamic patterns, providing a well-defined setting for evaluating how faithfully models can express visual information and spatio-temporal relationships through executable code.

\begin{figure}[t]
    \centering
    \includegraphics[width=0.85\linewidth]{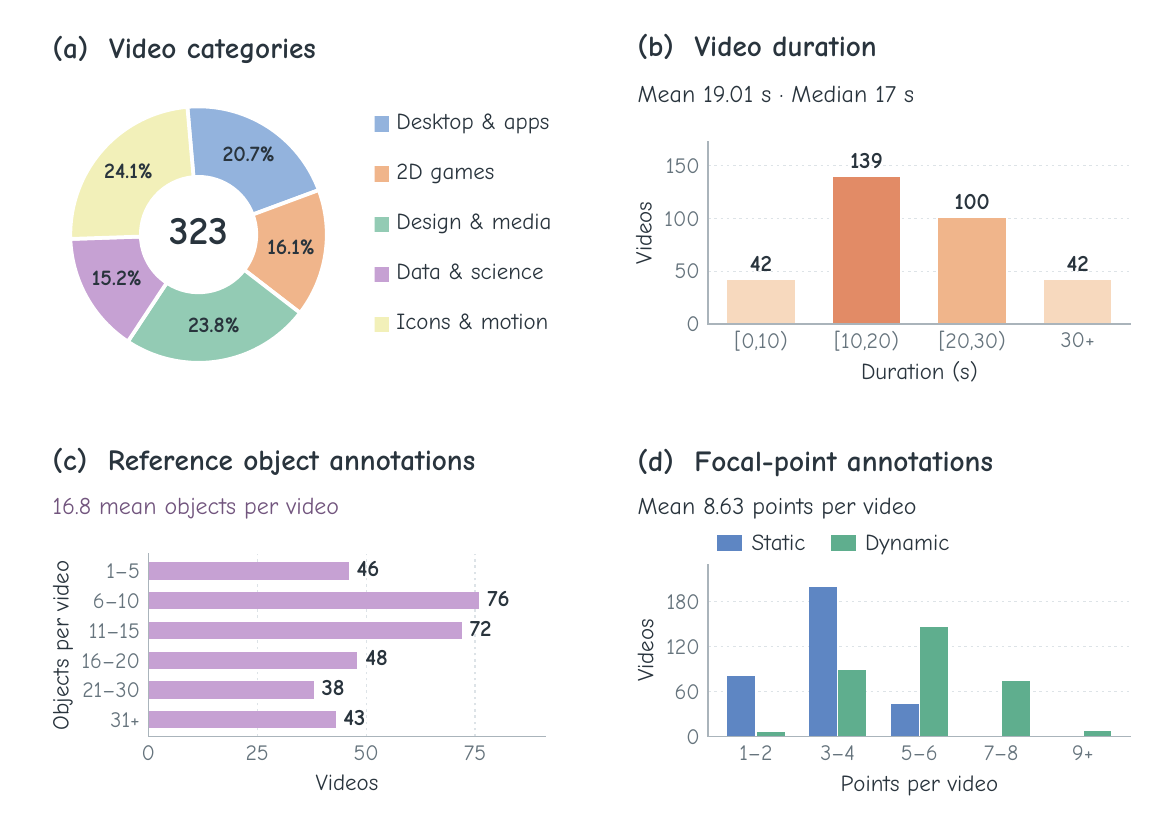}
    \caption{\textbf{Statistics of \name.} (a) Video counts and proportions across five categories. (b) Video duration distribution. (c) Reference objects per video. (d) Static and dynamic focal points per video.}
    \label{fig:codecap_dataset_statistics}
\end{figure}

\paragraph{Data collection.}
We collect candidate videos from public platforms, including Bilibili, YouTube, Behance, and Vimeo, covering five categories: desktop and application interfaces, 2D games, visual design and media animation, data and scientific visualization, and icons and abstract motion graphics.
These categories encompass diverse visual content and dynamic patterns, including window operations, page scrolling, object motion, dynamic charts, geometric construction, and shape transformations.

\paragraph{Content filtering.}
Candidate videos are filtered by scene geometry, appearance complexity, and motion structure.
We retain videos dominated by 2D layouts, shapes, text, and explicit object trajectories, while excluding those dominated by complex 3D viewpoints, detailed textures, or dense pixel-level dynamics.
This selection keeps the benchmark focused on visual content amenable to programmatic representation.

\paragraph{Annotation and human review.}
We provide two types of annotations: focal-point annotations and object-level spatio-temporal annotations.
Each focal-point annotation identifies a key static feature or dynamic process, with a description and corresponding timestamps in the reference video.
Static focal points capture the appearance, composition, spatial relationships, and text presentation of visual elements, while dynamic focal points capture state changes, motion processes, rhythm, and temporal coordination among objects.
Object-level spatio-temporal annotations record the visible intervals and per-frame bounding boxes of key objects.
We obtain spatial annotations through video segmentation model guided by bounding-box prompts together with positive/negative point prompts on representative frames.
All annotations undergo human review and correction.
Reviewers assess whether focal points are representative, accurately described, and correctly localized in time.
Object-level annotations are checked for consistent object identities, correct visible intervals, and accurate spatial boundaries.

\subsection{Dataset Statistics}
\label{subsec:dataset_statistics}

Figure~\ref{fig:codecap_dataset_statistics}(a--b) shows the distributions of video categories and durations.
\name comprises $323$ videos across five categories of structured 2D content.
The videos have a mean duration of $19.0$ seconds and a median duration of $17.0$ seconds.
In comparison, Animation2Code~\citep{ji2026animation2code} focuses on CodePen web animations with an average duration of $5.1$ seconds.
\name covers a broader range of structured 2D content and longer temporal sequences, allowing evaluation across more diverse visual structures and dynamic patterns.
Figure~\ref{fig:codecap_dataset_statistics}(c--d) shows the distributions of object and focal-point counts per video.
Each video contains an average of $16.8$ annotated objects and $8.6$ static and dynamic focal points.
The annotated focal-point intervals have a mean duration of $3.3$ seconds and a median of $1.8$ seconds.

\section{Experiments}
\label{sec:experiments}

Our experiments aim to understand how faithfully current MLLMs can express video content through executable code, particularly in preserving fine-grained visual details, spatial structures, and temporal dynamics.
We further investigate how agentic workflows and reasoning effort affect code captioning fidelity, and examine the trade-offs among visual fidelity, generation reliability, and computational cost.

\subsection{Experimental Setup}
\label{subsec:experimental_setup}

\begin{table*}[t]
    \centering
    \caption{\textbf{Main results on \name.} All configurations use medium thinking effort. \textbf{Bold} values indicate the \textbf{best} results within the Agentic and Non-agentic groups, respectively.}
    \label{tab:main_results}
    \footnotesize
    \setlength{\tabcolsep}{4pt}
    \renewcommand{\arraystretch}{1.15}
    \resizebox{\textwidth}{!}{
    \begin{tabular}{llcccccc}
        \toprule
        \multirow{2}{*}{\textbf{Model}} & \multirow{2}{*}{\textbf{Harness}} & \multirow{2}{*}{\makecell{\textbf{Video Production}\\\textbf{Success Rate} (\%) $\uparrow$}} & \multirow{2}{*}{\makecell{\textbf{Visual}\\\textbf{similarity} $\uparrow$}} & \multirow{2}{*}{\makecell{\textbf{Object-level}\\\textbf{Spatio-temporal}\\\textbf{IoU} $\uparrow$}} & \multicolumn{3}{c}{\makecell{\textbf{Fine-grained Visual}\\\textbf{Information Retention}}} \\
        \cmidrule(lr){6-8}
        & & & & & \textbf{Dynamic} $\uparrow$ & \textbf{Static} $\uparrow$ & \textbf{Overall} $\uparrow$ \\
        \midrule

        \rowcolor[gray]{0.93}
        \multicolumn{8}{c}{\textbf{Agentic}} \\
        \addlinespace[2pt]
        GPT-6-astra         & Codex       & 99.4 & 0.9735 & \textbf{75.37} & \textbf{68.61} & 80.72 & \textbf{73.24} \\
        GPT-6-astra         & Pi Agent    & 99.7 & \textbf{0.9771} & 74.45 & 67.66 & \textbf{81.35} & 72.87 \\
        Opus-5              & Pi Agent    & 97.8 & 0.9449 & 61.45 & 55.49 & 65.10 & 58.84 \\
        Opus-5              & Claude Code & 99.4 & 0.9555 & 58.55 & 53.23 & 63.40 & 56.88 \\
        GPT-6-sol           & Codex       & \textbf{100.0} & 0.9563 & 55.63 & 43.92 & 64.04 & 51.34 \\
        Qwen3.8-max         & Pi Agent    & 93.2 & 0.8869 & 48.90 & 44.81 & 55.88 & 48.83 \\
        Gemini-3.8-flash    & Pi Agent    & 98.8 & 0.9257 & 42.75 & 43.66 & 55.18 & 47.73 \\
        GPT-6-sol           & Pi Agent    & 99.1 & 0.9423 & 50.42 & 40.33 & 59.88 & 47.53 \\
        DeepSeek-v4.1-flash & Pi Agent    & 91.6 & 0.8692 & 45.14 & 40.45 & 51.27 & 44.29 \\
        GPT-5.6-sol         & Pi Agent    & 99.1 & 0.9320 & 46.10 & 38.60 & 53.36 & 44.02 \\
        GPT-5.6-sol         & Codex       & 99.7 & 0.9336 & 42.90 & 37.67 & 51.85 & 42.98 \\
        \addlinespace[4pt]

        \rowcolor[gray]{0.93}
        \multicolumn{8}{c}{\textbf{Non-agentic}} \\
        \addlinespace[2pt]
        GPT-6-astra         & -           & \textbf{98.5} & \textbf{0.9622} & \textbf{69.96} & \textbf{69.00} & \textbf{76.85} & \textbf{72.00} \\
        GPT-6-sol           & -           & 95.7 & 0.9168 & 58.01 & 48.54 & 61.07 & 53.08 \\
        GPT-5.6-sol         & -           & 96.9 & 0.9223 & 49.88 & 46.50 & 56.03 & 50.05 \\
        Gemini-3.8-flash    & -           & 97.2 & 0.9026 & 35.56 & 43.35 & 49.68 & 45.57 \\
        Qwen3.8-max         & -           & 82.0 & 0.7754 & 42.41 & 43.39 & 47.15 & 44.67 \\
        \bottomrule
    \end{tabular}
    }
\end{table*}

\textbf{Models and generation settings.}
We evaluate $7$ MLLMs: GPT-6-astra~\citep{openai2026gpt6astra}, GPT-6-sol~\citep{openai2026gpt6sol}, GPT-5.6-sol~\citep{openai2026gpt56sol}, Qwen3.8-max~\citep{qwen2026qwen38max}, DeepSeek-v4.1-flash~\citep{deepseek2026v41flash}, Opus-5~\citep{anthropic2026opus5}, and Gemini-3.8-flash~\citep{google2026gemini38flash}.
In the agentic setting, we evaluate models using Pi Agent, Codex, or Claude Code as the agent harness.
Models can inspect reference frames through tool calls, generate executable code, and iteratively refine their implementations using rendering feedback.
In the non-agentic setting, models generate code in a single pass from sampled video frames and the task prompt, without tool calls or rendering feedback.

\textbf{Input format.}
In the agentic setting, models access the reference video through tools and select timestamps for frame inspection.
In the non-agentic setting, reference videos are uniformly sampled at 2 fps.
When sampled frames exceeds the 50-image API limit for GPT models, we arrange them into temporally ordered image grids to retain all sampled frames.
Other models receive the sampled frames directly.

\textbf{Evaluation metrics.}
Following Section~\ref{subsec:evaluation_framework}, we evaluate code caption fidelity using visual similarity, object-level spatio-temporal IoU, and fine-grained visual information retention.
We further report static and dynamic retention scores to distinguish the preservation of visual appearance and spatial structure from that of motion and temporal relationships.
We additionally report the video production success rate, defined as the percentage of tasks that successfully produce a rendered video.
Tasks that fail to produce a video receive zero scores for all metrics.

\subsection{Main Results}
\label{subsec:main_results}

Table~\ref{tab:main_results} presents the performance of different models and generation configurations at medium thinking effort.
GPT-6-astra achieves the highest overall fine-grained visual information retention in both agentic and non-agentic settings.
With Codex, it reaches $73.24$ in overall retention and $75.37$ in object-level spatio-temporal IoU, outperforming the next-best model, Opus-5 with Pi Agent, by $14.40$ points in overall retention.
Other models exhibit substantial performance differences despite similarly high video production success rates.
For example, with Codex, GPT-6-sol achieves a 100\% success rate but only 51.34 in overall retention, compared with 73.24 for GPT-6-astra at a 99.4\% success rate.
These results indicate that successful video production does not necessarily imply faithful preservation of visual information.

Across all evaluated configurations, dynamic visual information is consistently less faithfully preserved than static content.
For example, GPT-6-astra with Codex achieves 80.72 in static retention but only 68.61 in dynamic retention, while GPT-6-sol with Codex shows a larger gap between 64.04 and 43.92.
This pattern holds across different models and generation settings, highlighting the difficulty of accurately expressing temporal dynamics through executable code.

Agentic generation consistently improves video production success rates but does not always improve visual fidelity.
Compared with non-agentic generation, Qwen3.8-max with Pi Agent improves its success rate from 82.0\% to 93.2\% and overall retention from 44.67 to 48.83.
In contrast, GPT-6-sol and GPT-5.6-sol achieve lower overall retention with both tested agent harnesses than in the non-agentic setting.
These results suggest that the benefits of agentic generation for code captioning fidelity depend on the model and generation configuration.

\begin{figure*}[t]
    \centering
    \includegraphics[width=\textwidth]{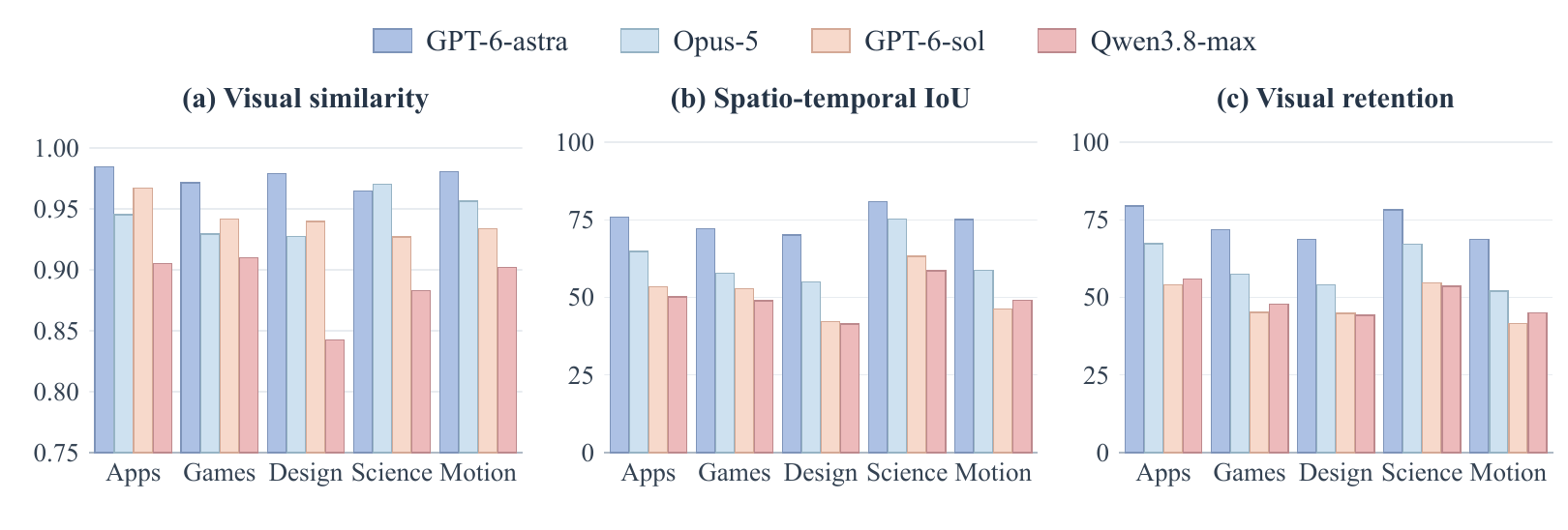}
    \caption{\textbf{Average scores of four models across video categories.} Across all four models, overall visual information retention is higher for desktop and application interfaces and data and scientific visualization than for visual design and media animation and icons and abstract motion graphics.}
    \label{fig:category_results}
\end{figure*}

Figure~\ref{fig:category_results} compares code captioning performance across five video categories for four models using Pi Agent at medium thinking effort.
GPT-6-astra leads in every category on both object-level spatio-temporal IoU and overall fine-grained visual information retention.
All four models achieve their highest object-level spatio-temporal IoU in the ``Data and Scientific Visualization'' category.
For overall retention, all four models perform better on ``Desktop and Application Interfaces'' and ``Data and Scientific Visualization'' than on either ``Visual Design and Media Animation'' or ``Icons and Abstract Motion Graphics.''
These results highlight differences in code captioning fidelity across video categories.

\subsection{Further Analysis}
\label{subsec:further_analysis}

\subsubsection{Effect of Thinking Effort}
\label{subsubsec:thinking_scaling}

\begin{figure}[t]
    \centering
    \includegraphics[width=\linewidth]{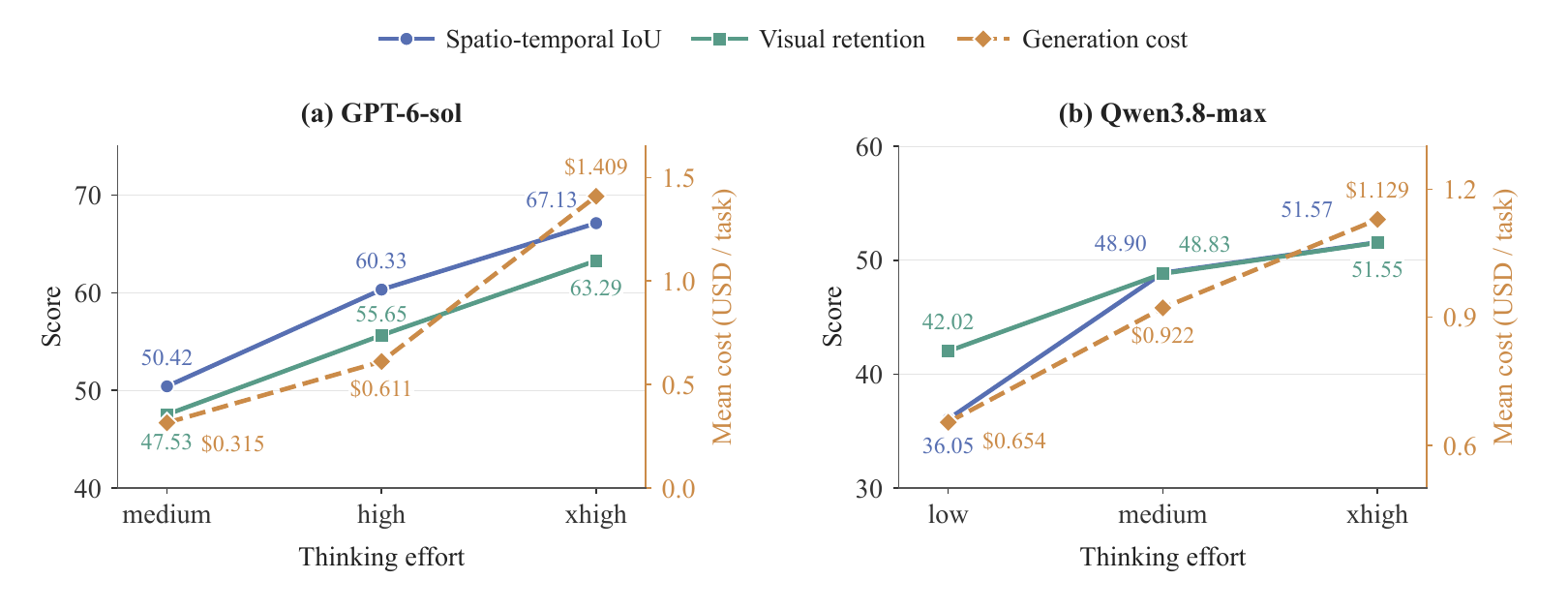}
    \caption{\textbf{Visual fidelity and generation cost across different thinking effort levels using the Pi Agent harness.} (a) GPT-6-sol; (b) Qwen3.8-max. The left axis shows object-level spatio-temporal IoU and fine-grained visual information retention, while the right axis shows average generation cost.}
    \label{fig:thinking_scaling}
\end{figure}

As shown in Figure~\ref{fig:thinking_scaling}, increasing thinking effort improves code captioning fidelity for both models, but the magnitude of improvement and the associated generation cost differ.
For GPT-6-sol, increasing thinking effort from medium to xhigh raises object-level spatio-temporal IoU from 50.42 to 67.13 and fine-grained visual information retention from 47.53 to 63.29.
Meanwhile, the average generation cost increases from USD 0.315 to USD 1.409, indicating that substantial fidelity gains come with considerably higher costs.

For Qwen3.8-max, most improvements occur when thinking effort increases from low to medium, with spatio-temporal IoU rising from 36.05 to 48.90 and visual information retention from 42.02 to 48.83.
Further increasing effort to xhigh yields smaller gains of 2.67 and 2.72 points, respectively, while the average generation cost increases from USD 0.922 to USD 1.129.
These results reveal model-dependent returns to increased thinking effort, with GPT-6-sol maintaining larger fidelity gains at higher effort levels than Qwen3.8-max.

\subsubsection{Generation Cost and Efficiency}
\label{subsubsec:cost_efficiency}

\begin{figure}[t]
    \centering
    \includegraphics[width=\linewidth]{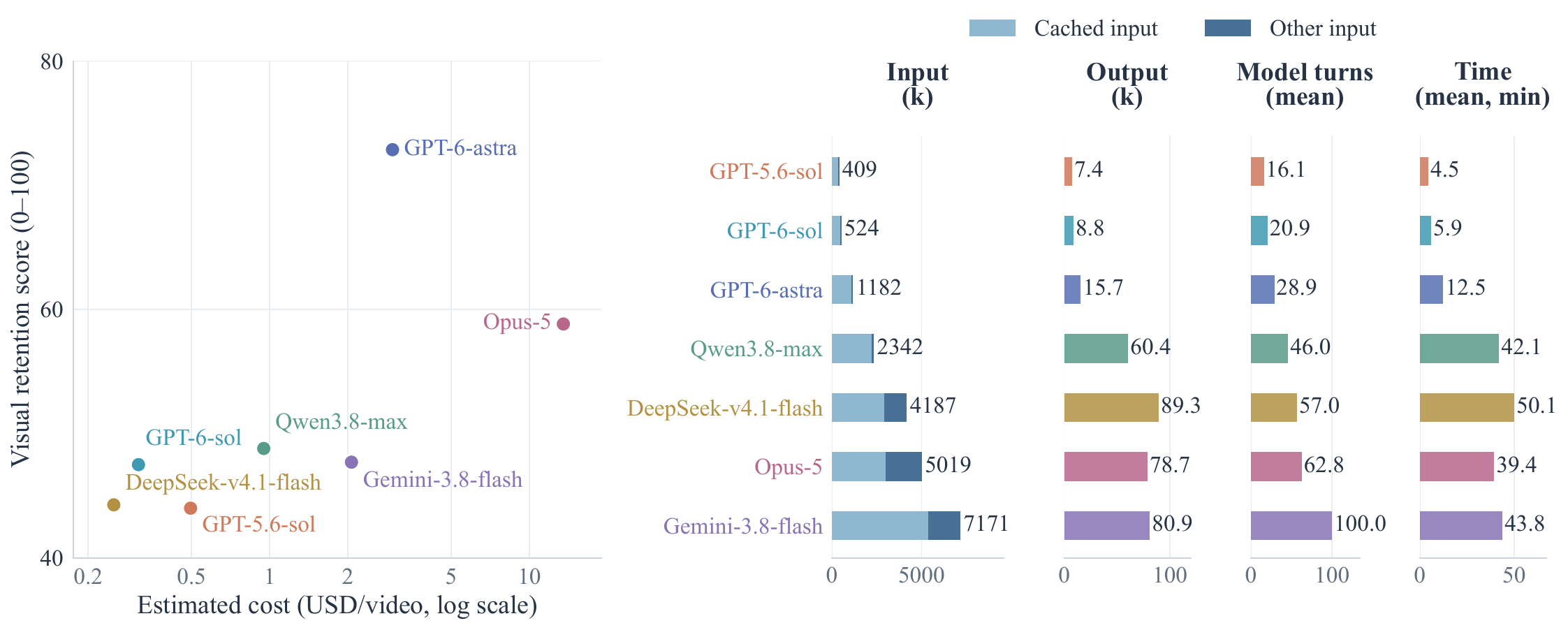}
    \caption{Generation cost and efficiency with the Pi Agent harness.}
    \label{fig:cost_efficiency}
\end{figure}

We examine the relationship between code captioning fidelity, generation cost, and runtime efficiency across models using the Pi Agent harness.
As shown in Figure~\ref{fig:cost_efficiency}, higher generation cost does not consistently translate into better caption fidelity.
For example, GPT-6-astra achieves an overall visual information retention score of 72.87 at a substantially lower estimated cost than Opus-5, which scores 58.84.
Qwen3.8-max and DeepSeek-v4.1-flash also achieve comparable or higher retention scores than some other models at lower estimated costs, indicating that monetary cost alone does not determine captioning performance.

However, lower generation cost does not necessarily imply greater runtime efficiency.
Qwen3.8-max and DeepSeek-v4.1-flash take an average of 42.1 and 50.1 minutes per video, respectively, compared with 12.5 minutes for GPT-6-astra.
They also consume more tokens and involve more model response rounds than the GPT series.
These results highlight the differences between monetary cost, token consumption, and generation time when evaluating the efficiency of code captioning systems.

\begin{figure}[t]
    \centering
    \includegraphics[width=1\linewidth]{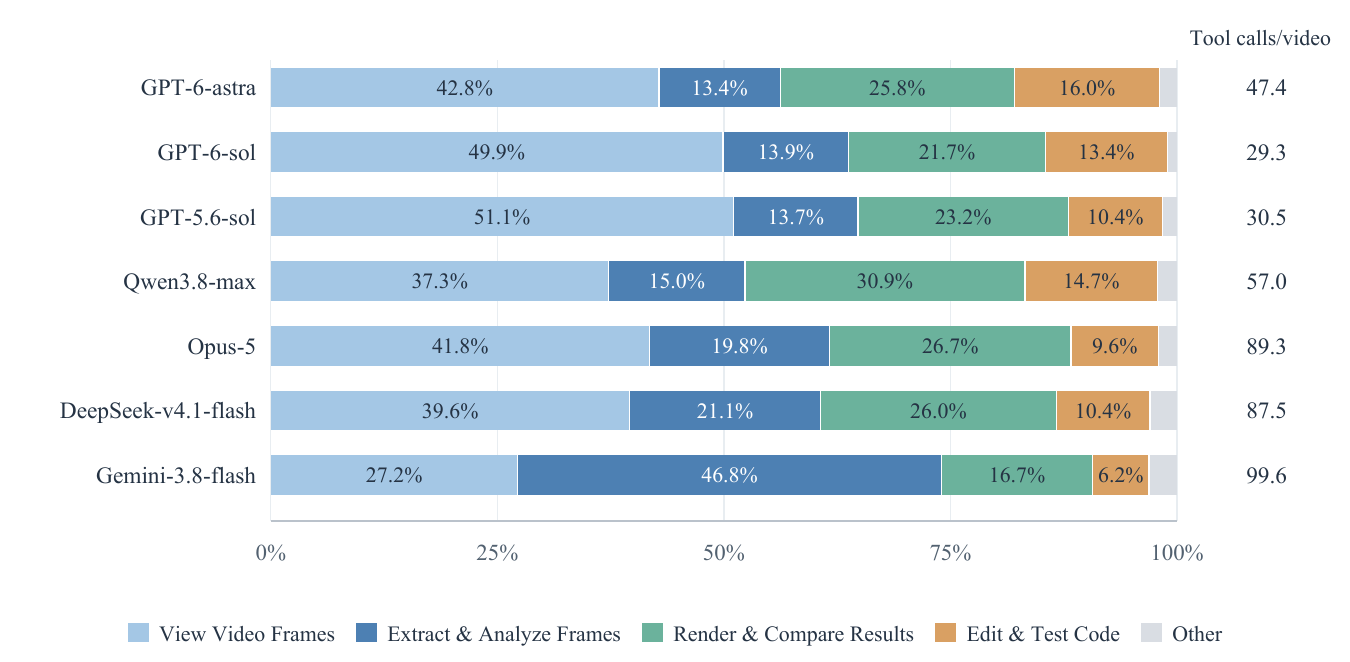}
    \caption{Tool call composition for seven models with the Pi Agent harness.}
    \label{fig:tool_usage}
\end{figure}

\subsubsection{Tool Usage Behavior}
\label{subsubsec:tool_usage}

Figure~\ref{fig:tool_usage} compares the tool usage patterns of seven models under the same Pi Agent harness.
GPT-6-sol and GPT-5.6-sol devote approximately half of their tool calls to directly viewing reference frames, while Gemini-3.8-flash allocates 46.8\% of its calls to extracting and analyzing frames.
Qwen3.8-max has the highest proportion of rendering and comparison calls at 30.9\%, indicating greater emphasis on inspecting generated results.
These differences show that models distribute their tool calls differently across reference observation, frame analysis, and rendering feedback, even under the same agent harness.

The total number of tool calls also varies considerably across models, but more calls do not necessarily correspond to higher caption fidelity.
For example, GPT-6-astra achieves an overall visual information retention score of 72.87 with an average of 47.4 tool calls per video, whereas Gemini-3.8-flash scores 47.73 with 99.6 calls.
These results suggest that tool usage frequency alone is insufficient to characterize the effectiveness of agentic code captioning.

\subsubsection{Validity of the Evaluation Metric}
\label{subsubsec:metric_validity}

\begin{figure*}[t]
    \centering
    \begingroup
    \sbox{0}{%
        \small
        \setlength{\tabcolsep}{4pt}
        \renewcommand{\arraystretch}{1.35}
\begin{tabular}{llcc}
            \toprule
            \multirow{2}{*}{\textbf{Model}} & \multirow{2}{*}{\textbf{Metric}} & \multicolumn{2}{c}{\textbf{Judge model}} \\
            \cmidrule(lr){3-4}
            & & GPT-6-sol & Gemini-3.1-pro \\
            \midrule
            \multirow{3}{*}{GPT-6-sol}
                & Dynamic & 40.33 & 35.73 \\
                & Static & 59.88 & 59.55 \\
                & \textbf{Overall} & \textbf{47.53} & \textbf{44.39} \\
            \midrule
            \multirow{3}{*}{\makecell{DeepSeek-\\v4.1-flash}}
                & Dynamic & 40.45 & 35.30 \\
                & Static & 51.27 & 50.18 \\
                & \textbf{Overall} & \textbf{44.29} & \textbf{40.81} \\
            \bottomrule
        \end{tabular}
    }
    \begin{minipage}[t]{0.46\textwidth}
        \vspace{0pt}\centering
        \resizebox{0.96\linewidth}{!}{\usebox{0}}
    \end{minipage}\hfill
    \begin{minipage}[t]{0.50\textwidth}
        \vspace{0pt}\centering
        \includegraphics[height=\dimexpr\ht0+\dp0\relax]{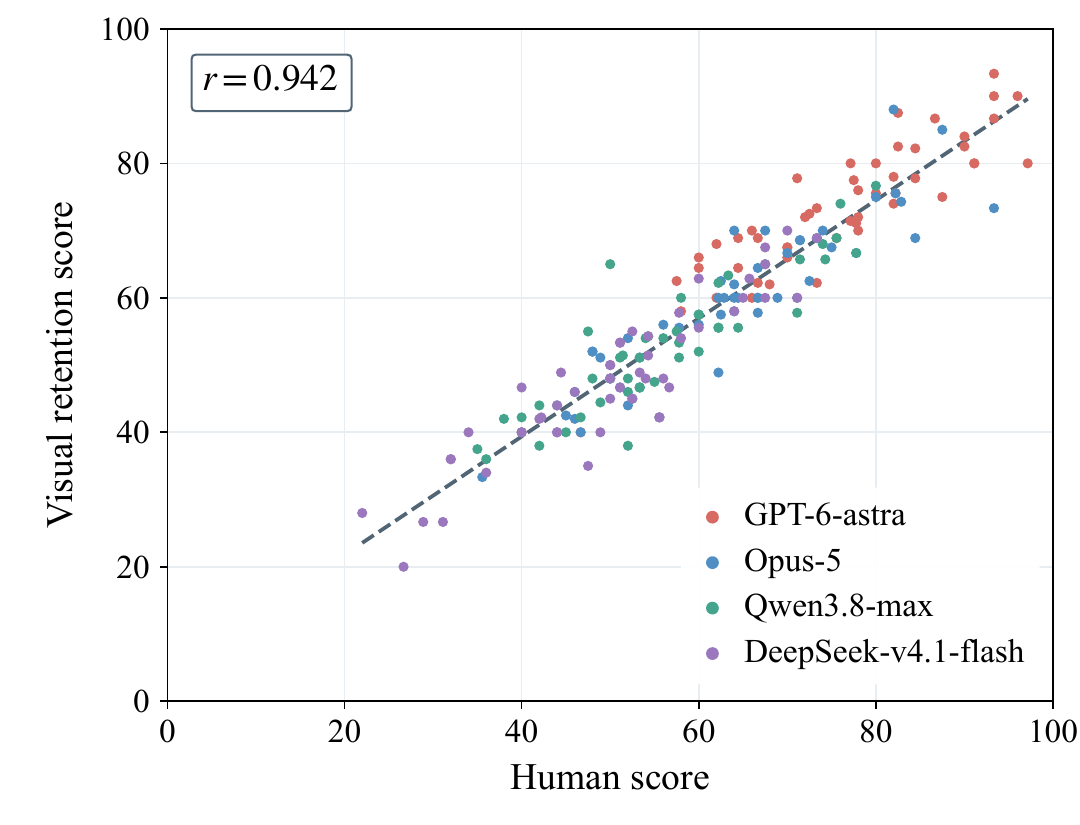}
    \end{minipage}
    \par\medskip
    \begin{minipage}[t]{0.46\textwidth}
        \centering\small (a) Scores by judge model
    \end{minipage}\hfill
    \begin{minipage}[t]{0.50\textwidth}
        \centering\small (b) Correlation between MLLM and human scores
    \end{minipage}
    \endgroup
    \caption{\textbf{Scores from different judge models and their correlation with human scores}. The two judge configurations yield the same overall ranking for the two evaluated generation models, and GPT-6-sol judge scores correlate strongly with human scores (Pearson r=0.942).}
    \label{fig:human_alignment}
\end{figure*}

Figure~\ref{fig:human_alignment}(a) compares fine-grained visual information retention scores assigned by GPT-6-sol and Gemini-3.1-pro-preview to outputs generated by GPT-6-sol and DeepSeek-v4.1-flash using Pi Agent at medium thinking effort.
Both judges use the same focal-point annotations, scoring criteria, and temporal sampling rate of 2 fps.
GPT receives sampled frames as individual images or image grids, while Gemini processes video inputs at their native spatial resolution.
Although their absolute scores differ, both judges assign higher overall retention scores to GPT-6-sol than to DeepSeek-v4.1-flash, showing consistent relative assessments for the two evaluated models.

To assess alignment with human judgments, we sample 50 reference videos, with 10 from each category, and evaluate outputs generated by GPT-6-astra, Opus-5, Qwen3.8-max, and DeepSeek-v4.1-flash using Pi Agent.
Human annotators compare the generated videos with the corresponding reference segments and score each focal point using the same 0--5 criteria as our evaluation metric.
We aggregate focal-point scores into video-level scores and compute the Pearson correlation between human and GPT-6-sol judge scores for the same generated outputs.
As shown in Figure~\ref{fig:human_alignment}(b), the Pearson correlation coefficient reaches 0.942, indicating strong alignment between MLLM-based evaluation and human assessments of visual information retention.

\subsubsection{Execution Trace Example}
\label{subsubsec:execution_example}

\begin{figure*}[t]
    \centering
    \includegraphics[width=\textwidth]{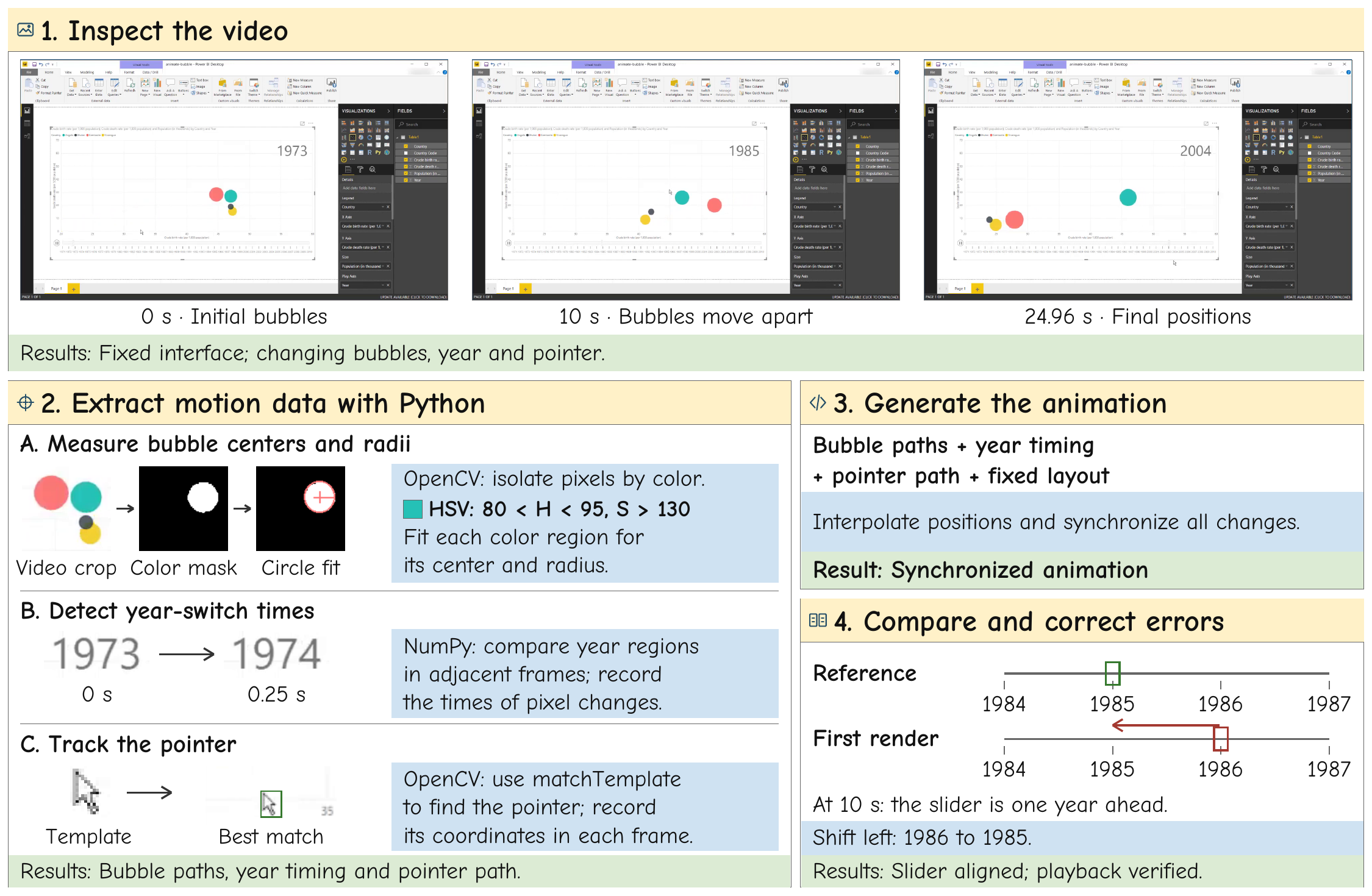}
    \caption{Execution trace of GPT-6-astra reconstructing a bubble chart video with the Codex harness.}
    \label{fig:execution_example}
\end{figure*}

Figure~\ref{fig:execution_example} illustrates how GPT-6-astra uses the Codex harness to generate and refine a code caption for a bubble chart video.
The model first separates static interface elements from dynamic objects and uses color segmentation to estimate bubble centers and radii.
It then detects year transitions through frame differences and tracks the pointer using template matching.
These measurements are combined with the fixed layout to generate the animation.
By comparing the rendered and reference frames at the same timestamp, the model identifies a one-year offset in the slider and corrects its position.
This example illustrates how tool-assisted visual analysis extracts structured spatio-temporal information for code generation, while rendering feedback helps detect and correct temporal misalignment.

\subsubsection{Comparing Code and Natural Language Captions}
\label{subsubsec:code_vs_nl}

\paragraph{Information Retention for Question Answering.}
We randomly select 100 videos with question--answer annotations from DVBench~\citep{wang2026dvbench}, VideoMathQA~\citep{rasheed2026videomathqa}, and GUI-World~\citep{chen2025gui} to compare the information retained by code and natural language captions.
For each video, GPT-5.6-sol, Opus-5, and Qwen3.8-max generate both types of captions.
Gemini-3.8-flash answers the annotated questions using either the video rendered from the code caption or the natural language caption, and its predictions are evaluated against ground-truth answers.
We also report its accuracy when answering directly from the original videos as a reference.

\begin{table}[htbp]
    \centering
    \caption{Question-answering accuracy (\%) from rendered videos and caption text.}
    \label{tab:qa_retention}
    \small
    \setlength{\tabcolsep}{5pt}
    \renewcommand{\arraystretch}{1.15}
    \begin{tabular}{lcc}
        \toprule
        \textbf{Caption model}
        & \makecell{\textbf{Code-rendered}\\\textbf{video QA} $\uparrow$}
        & \makecell{\textbf{Natural language}\\\textbf{caption QA} $\uparrow$} \\
        \midrule
        GPT-5.6-sol & 76.42 & 73.58 \\
        Opus-5 & 83.02 & 72.64 \\
        Qwen3.8-max & 81.13 & 78.30 \\
        \midrule
        Original videos & \multicolumn{2}{c}{83.96} \\
        \bottomrule
    \end{tabular}
\end{table}

As shown in Table~\ref{tab:qa_retention}, code-rendered videos achieve higher question-answering accuracy than natural language captions across all three caption models.
The improvements range from 2.83 to 10.38 percentage points, with Opus-5 showing the largest gain.
Its code-rendered videos achieve 83.02\% accuracy, approaching the 83.96\% obtained from the original videos.
These results suggest that, under the evaluated setting, code captions preserve more information useful for answering the annotated questions.

\paragraph{Video Reconstruction Fidelity.}
We further compare the reconstruction fidelity of code and natural language captions on \name.
Code captions are directly rendered into videos, while natural language captions are converted into videos using Seedance-2.5~\citep{bytedance2026seedance25}.
We evaluate reconstruction fidelity using Appearance Similarity (AS), Motion Pattern Similarity (MPS), and Keyframe Structural Similarity (KSS).
Each reference video is divided into shots, which are matched with corresponding segments in the generated video.
AS measures DreamSim~\citep{fu2023dreamsim} feature similarity, MPS compares the directions and magnitudes of point displacements tracked by CoTracker3~\citep{karaev2025cotracker3}, and KSS computes SSIM~\citep{wang2004image} between each reference shot's midpoint frame and its matched generated frame.
Scores are averaged across shots within each video and then across videos.

\begin{table}[htbp]
    \centering
    \caption{Video reconstruction fidelity from code and natural language captions.}
    \label{tab:reconstruction_comparison}
    \small
    \setlength{\tabcolsep}{4pt}
    \renewcommand{\arraystretch}{1.15}
    \begin{tabular}{llccc}
        \toprule
        \textbf{Caption model} & \textbf{Method}
        & \textbf{AS} $\uparrow$
        & \textbf{MPS} $\uparrow$
        & \textbf{KSS} $\uparrow$ \\
        \midrule
        \multirow{2}{*}{GPT-5.6-sol}
            & Code rendering
            & \textbf{0.8907} & \textbf{0.5412} & \textbf{0.8068} \\
            & Caption-to-video
            & 0.6628 & 0.5156 & 0.5925 \\
        \midrule
        \multirow{2}{*}{Opus-5}
            & Code rendering
            & \textbf{0.9148} & \textbf{0.5572} & \textbf{0.8165} \\
            & Caption-to-video
            & 0.6932 & 0.5150 & 0.5958 \\
        \midrule
        \multirow{2}{*}{Qwen3.8-max}
            & Code rendering
            & \textbf{0.8773} & \textbf{0.5488} & \textbf{0.7817} \\
            & Caption-to-video
            & 0.6754 & 0.5185 & 0.5984 \\
        \bottomrule
    \end{tabular}
\end{table}

Table~\ref{tab:reconstruction_comparison} shows that code rendering consistently achieves higher scores across all three metrics and caption models.
The gains are particularly pronounced in appearance and structural similarity, with AS increasing by 0.20--0.23 and KSS by 0.18--0.22.
Motion pattern similarity also improves, although the numerical gains are smaller.
Unlike natural language captions, which require a separate generative model to produce visual content, executable code explicitly specifies how visual elements are constructed and evolve over time.
These results highlight the potential of code captions to preserve fine-grained visual information and spatio-temporal structure through an explicit and executable representation.

\section{Conclusion}
\label{sec:conclusion}

In this work, we explore the potential of executable code as a complementary representation for video captioning and introduce \textbf{\name} to systematically evaluate the code captioning capabilities of MLLMs.
Focusing on structured 2D videos, \name comprises diverse video samples, fine-grained annotations, and a multi-level evaluation framework for assessing visual and spatio-temporal fidelity.
Our experiments show that executable code better preserves visual information than natural language captions in question answering and video reconstruction. 
These results suggest that code can serve as a promising complementary representation for video captioning.
Meanwhile, evaluations on \name reveal that leading proprietary MLLMs demonstrate promising code captioning capabilities but still struggle to capture complex temporal dynamics, while open-source models exhibit substantial performance gaps.
We further find that agentic workflows improve rendering success rates but do not consistently enhance caption fidelity.
Together, these findings highlight both the promise of executable code for video captioning and the challenges of generating faithful code captions, providing a foundation for future research on structured video representations.

\clearpage
\bibliographystyle{plainnat}
\bibliography{references}

\clearpage
\section*{Appendix}
\appendix

\section{Evaluation Environment and Runtime Configuration}
\label{app:environment}

Table~\ref{tab:appendix_agent_environment} lists the harness versions and time limits used in the agentic code generation experiments. Each task runs in an isolated container, reads the reference video, generates a single-file web page, and renders it into a video through a browser. Separate time limits apply to code generation and video rendering.

\begin{table*}[htbp]
    \centering
    \caption{Runtime configuration for the agentic code generation experiments.}
    \label{tab:appendix_agent_environment}
    \small
    \setlength{\tabcolsep}{7pt}
    \renewcommand{\arraystretch}{1.16}
    \begin{tabular}{p{0.16\textwidth}p{0.30\textwidth}p{0.46\textwidth}}
        \toprule
        Item & Configuration & Description \\
        \midrule
        Agent harness & Pi Agent 1.1.0; Codex CLI 0.155.1; Claude Code 2.1.296 & Versions used for code generation \\
        Code generation time limit & 3,600 seconds & Timed separately for each video; Qwen3.8-max at xhigh thinking effort has a 7,200-second limit \\
        Video rendering time limit & 900 seconds & Time limit for a single rendering process \\
        Video rendering frame rate & 24 fps & Call the page's \texttt{seek(seconds)} interface at $1/24$-second intervals along the video timeline, capture screenshots, and encode them into a video \\
        \bottomrule
    \end{tabular}
\end{table*}

\section{Analysis of Video Generation Failures}
\label{subsubsec:error_analysis}

\begin{figure}[htbp]
    \centering
    \includegraphics[width=\linewidth]{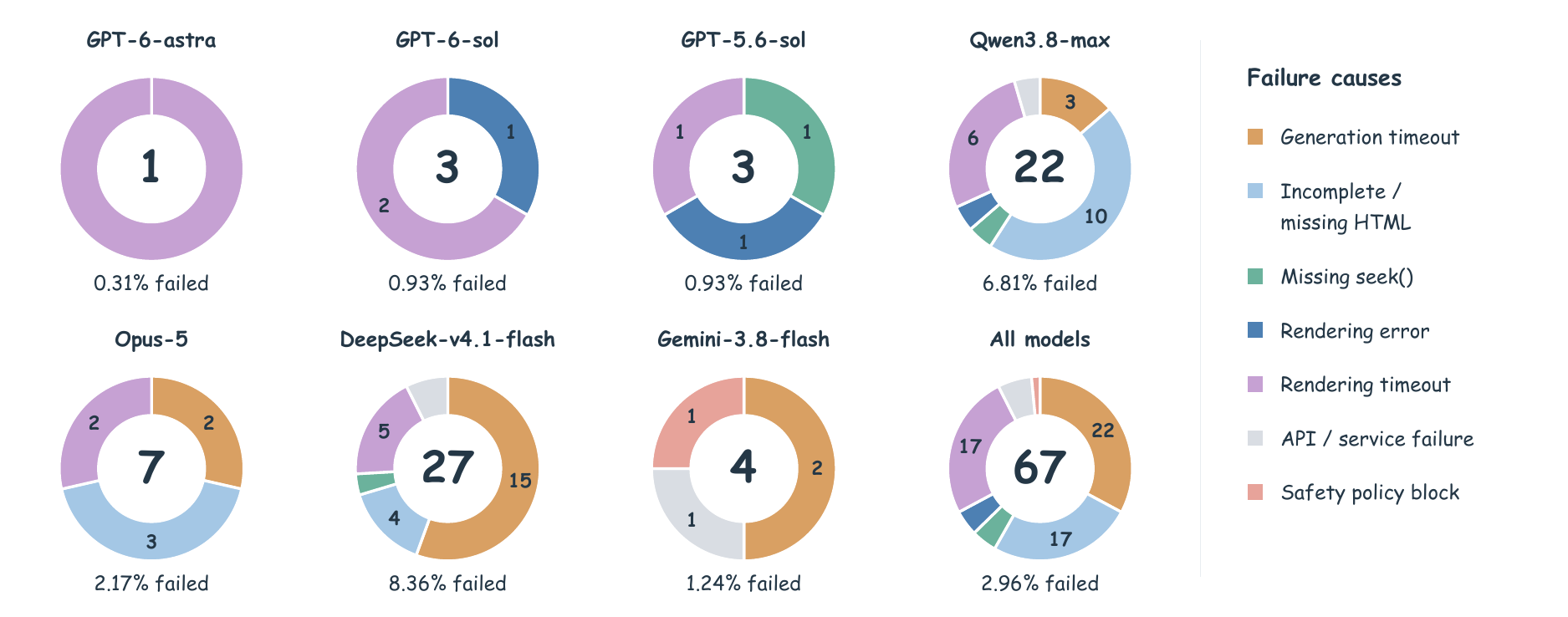}
    \caption{Video generation failures for seven models with Pi Agent. Sectors show the shares of failure causes, the center shows the failure count, and the label below shows the failure rate.}
    \label{fig:error_analysis}
\end{figure}

In Figure~\ref{fig:error_analysis}, generation timeouts, failure to deliver complete HTML, and rendering timeouts together account for 83.6\% of failed tasks. The main obstacles to producing a video occur at three stages: code generation, page delivery, and rendering.

Qwen3.8-max and DeepSeek-v4.1-flash together account for 73.1\% of all failed tasks, but their main failure causes differ. HTML delivery problems are the leading cause for Qwen3.8-max, accounting for 45.5\% of its failed tasks, while generation timeouts lead for DeepSeek-v4.1-flash at 55.6\%. All GPT models have failure rates below 1\%, and all their failures occur during rendering: rendering timeouts, runtime errors, or a missing interface for seeking to a specified time.

\section{Prompt Details}
\label{app:prompt}

This section lists the prompts used for agentic code generation, non-agentic code generation, and video scoring. The generation prompts specify requirements for implementing visuals entirely in code, aligning the timeline, and delivering a single file. The scoring prompt specifies how to compare the current focal point, the scoring criteria, and the output format. File paths and commands in the prompts reflect the actual configuration inside the evaluation container.

\subsection{Agentic Code Generation Prompt}
\begin{promptbox}
Reference video: /input/reference.mp4

Recreate the entire reference video, from the first frame to the last. The recreation must have the same total duration as the reference video.\par
The recreation starts at 0 seconds, and its timeline must align with the reference video's timeline. seek(seconds) takes a time on this shared timeline, from 0 through the video's total duration.\par
Use matching timestamps when extracting reference frames and inspecting the generated page.\par

\textbf{Task}\par
Recreate the visuals and all changes over time in the entire reference video using only HTML, CSS, and JavaScript. Preserve the original aspect ratio, main content, composition, and timing.

\textbf{Visual implementation constraints}\par
All visual content must be generated using DOM/CSS, Canvas drawing commands, inline SVG, or procedural WebGL. This includes backgrounds, text, icons, illustrations, software interfaces, and controls.\par
The reference video and extracted frames are for observation only and must not be used in the final rendering. Do not use video, full-frame or cropped screenshots, external images, AI-generated images, glyph atlases, extracted textures, or external fonts. Do not circumvent these restrictions by embedding reference imagery as Base64 data, pixel arrays, per-pixel drawing primitives, automatically traced shapes, or similar representations.\par
Render text from actual strings using the browser's text layout engine or Canvas fillText/strokeText. You may use fonts already installed on the system.\par
Rebuild interfaces, controls, illustrations, and textures in code, preserving their main content, composition, and colors. Do not simply omit them or replace them with unrelated placeholders.\par
Preserve the motion and transition timing of the reference video. Do not replace continuous motion with cuts between a few static images.\par

\textbf{Workflow and environment}\par
The working directory is /workspace. First extract reference frames yourself and inspect them with read, then write the code. Next, compare browser screenshots against the reference frames and iteratively refine the implementation, covering the main scenes and transitions. Do not infer the visuals from filenames alone.

The environment comes with ffprobe, ffmpeg, Python, Node, and Chromium preinstalled. Playwright is available at /opt/runtime/browser/node\_modules/playwright. Run Python scripts with the python command. Direct internet access is unavailable. Do not download or install additional dependencies.\par
Store reference frames for inspection, comparison screenshots, and validation scripts in /workspace/checks/. Store temporary files in /tmp.\par
Example screenshot command (at the start of both the recreation and the reference video): node /opt/tools/capture.mjs /workspace/index.html 0 /workspace/checks/frame.png\par
To inspect other times, replace 0 in the command with a specific timestamp in seconds, from 0 through the video's total duration. The command accepts a numeric value, not an arithmetic expression.\par

\textbf{Delivery requirements}\par
Save the entire implementation in /workspace/index.html, with no dependencies on other files or network resources. The page must play automatically when opened and scale to fit the window while preserving its aspect ratio.\par
Expose window.videoCompare = \{ duration, width, height, seek(seconds) \{ /* Pause playback and render the requested time. */ \} \}, where duration is the reference video's actual total duration in seconds, and width and height match its dimensions in pixels. seek(seconds) must pause the playback clock and reliably render the requested time, including when calls arrive out of chronological order or repeatedly target the same time.\par
The deliverable must run in an isolated environment containing only index.html, without reading checks/ or the reference video.\par
In your final response, state the entry point, how to open it, what validation you performed, and any remaining differences caused by approximations in the code-only recreation.\par
\end{promptbox}

\subsection{Non-Agentic Code Generation Prompt}
\begin{promptbox}
Reference input: the supplied frames sampled from the reference video and the accompanying video dimensions, total duration, and frame timestamps. If frames are arranged in collages, use the supplied layout and timestamp labels to identify each frame and its place in the timeline.

Recreate the entire reference video, from the first frame to the last. The recreation must have the same total duration as the reference video.\par
The recreation starts at 0 seconds, and its timeline must align with the reference video's timeline. seek(seconds) takes a time on this shared timeline, from 0 through the video's total duration.\par
Use the supplied timestamps to align the visual changes with the reference video's timeline.\par

\textbf{Task}\par
Recreate the visuals and all changes over time in the entire reference video using only HTML, CSS, and JavaScript. Preserve the original aspect ratio, main content, composition, and timing.

\textbf{Visual implementation constraints}\par
All visual content must be generated using DOM/CSS, Canvas drawing commands, inline SVG, or procedural WebGL. This includes backgrounds, text, icons, illustrations, software interfaces, and controls.\par
The reference video and extracted frames are for observation only and must not be used in the final rendering. Do not use video, full-frame or cropped screenshots, external images, AI-generated images, glyph atlases, extracted textures, or external fonts. Do not circumvent these restrictions by embedding reference imagery as Base64 data, pixel arrays, per-pixel drawing primitives, automatically traced shapes, or similar representations.\par
Render text from actual strings using the browser's text layout engine or Canvas fillText/strokeText. You may use fonts already installed on the system.\par
Rebuild interfaces, controls, illustrations, and textures in code, preserving their main content, composition, and colors. Do not simply omit them or replace them with unrelated placeholders.\par
Preserve the motion and transition timing of the reference video. Do not replace continuous motion with cuts between a few static images.\par

\textbf{Delivery requirements}\par
Return the entire implementation as the complete contents of a single index.html file, with no dependencies on other files or network resources.\par
Expose window.videoCompare = \{ duration, width, height, seek(seconds) \{ /* Pause playback and render the requested time. */ \} \}, where duration is the supplied reference video's actual total duration in seconds, and width and height match its supplied dimensions in pixels. seek(seconds) must pause the playback clock and reliably render the requested time, including when calls arrive out of chronological order or repeatedly target the same time.\par
The deliverable must run in an isolated environment containing only index.html, without reading the reference video or supplied images.\par
Output only the complete HTML document, starting with <!DOCTYPE html> and ending with </html>. Do not include Markdown code fences, explanations, or any text outside the HTML document.\par
\end{promptbox}

\subsection{Video Scoring Prompt}
\begin{promptbox}
Compare the generated video with the reference clip and rate how faithfully it reproduces the current focus on a scale from 0 to 5.

Look through the complete generated video to find the corresponding content. When similar content appears more than once, use the surrounding states to identify the matching sequence.\par
For static points, compare content, appearance, position, proportions, layout, spatial relationships, and information hierarchy. Check how long a state lasts when the focus requires it to persist.\par
For dynamic points, compare starting states, intermediate changes, order, pacing, duration, coordination, and results. When the focus requires other areas to stay stable, compare their behavior too.\par
Base the comparison on clearly visible frames and the details and relationships covered by the focus. The importance of a difference depends on what it changes, rather than how much screen space it occupies.\par

\textbf{Scores}\par
5: The content, appearance, structure, key relationships, and progression covered by the focus are accurate and complete. Timing correspondence is correct, and no difference is discernible in the visible frames.\par
4: The main content, key relationships, and complete progression are accurate. The overall match is very close, with only isolated, very slight differences in position, proportion, or local timing.\par
3: Core content, structure, and key relationships are complete and correct. For dynamics, the required key states, steps, feedback, coordination, and results are reproduced, with essentially correct main order and pacing. Non-core details have clear local deviations, while the overall presentation remains close to the reference.\par
2: There are clear omissions, errors, or widespread deviations overall, but at least one substantive part directly relevant to the focus is complete and accurate. A static part includes specific content and correct structural relationships; a dynamic part includes the corresponding starting state, coherent changes, and correct result.\par
1: Some relevant content or motion is reproduced, with rough or scattered correspondence. It does not yet form a complete, accurate, substantive local structure, relationship, or action process.\par
0: The focus is not substantively reproduced. For a purely dynamic point, a static end state without the corresponding action scores 0. When the focus includes other requirements, score the parts actually reproduced.\par

Timing and duration also affect the score. Appearing earlier or later as a whole, ending too early, or lasting too long all count as differences. An offset that preserves the main pacing and event relationships is minor; a clear change to a stage's timeline placement, the main pacing, or synchronization limits the score to 2.\par
Choose the highest score that matches the degree of reproduction achieved.\par

\textbf{Output}\par
Output only \{"score": N\}, where N is an integer from 0 to 5.

\textbf{Input}\par
\{Input\}\par
\end{promptbox}

\clearpage
\section{Representative Case Comparisons}
\label{app:representative_cases}

We select one case from each of the five dataset categories to compare the reference video with outputs generated by four models using Pi Agent. Each case shows four key timestamps, with identical timestamps and crop regions across rows. From top to bottom, the rows show the reference video, GPT-6-astra, Opus-5, Qwen3.8-max, and GPT-5.6-sol. Columns are arranged in chronological order.

\begin{figure*}[htbp]
    \centering
    \includegraphics[width=\textwidth]{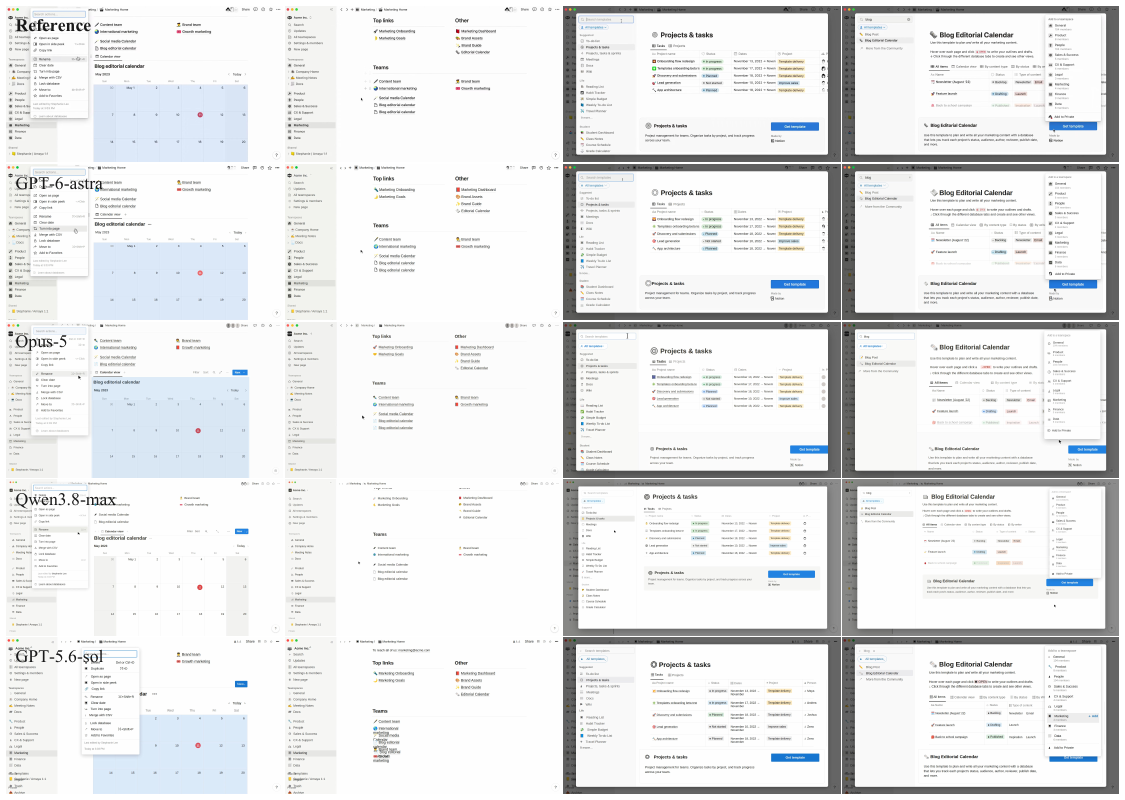}
    \caption{Workspace interface. The four columns correspond to 3.0, 5.0, 11.5, and 16.0 seconds, showing a calendar view, workspace navigation, and database template previews.}
    \label{fig:appendix_case_a}
\end{figure*}

\begin{figure*}[htbp]
    \centering
    \includegraphics[width=\textwidth]{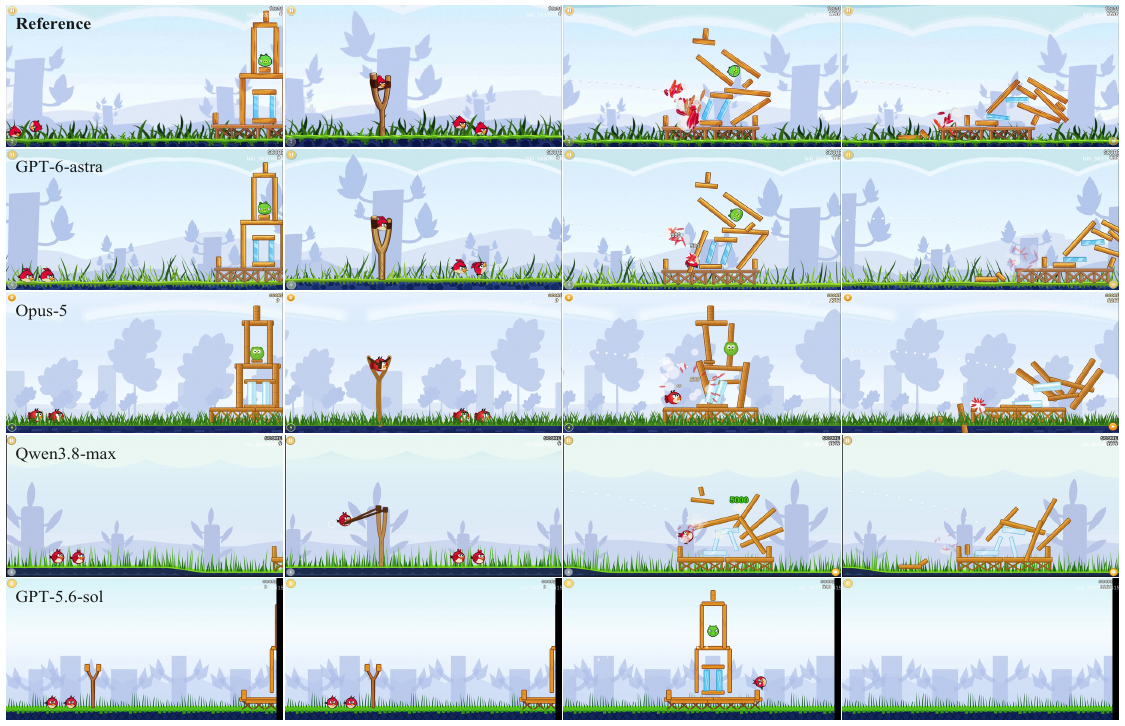}
    \caption{2D game. The four columns correspond to 3.1, 5.0, 8.0, and 13.4 seconds, showing the launch, collision, and structural collapse.}
    \label{fig:appendix_case_b}
\end{figure*}

\begin{figure*}[htbp]
    \centering
    \includegraphics[width=\textwidth]{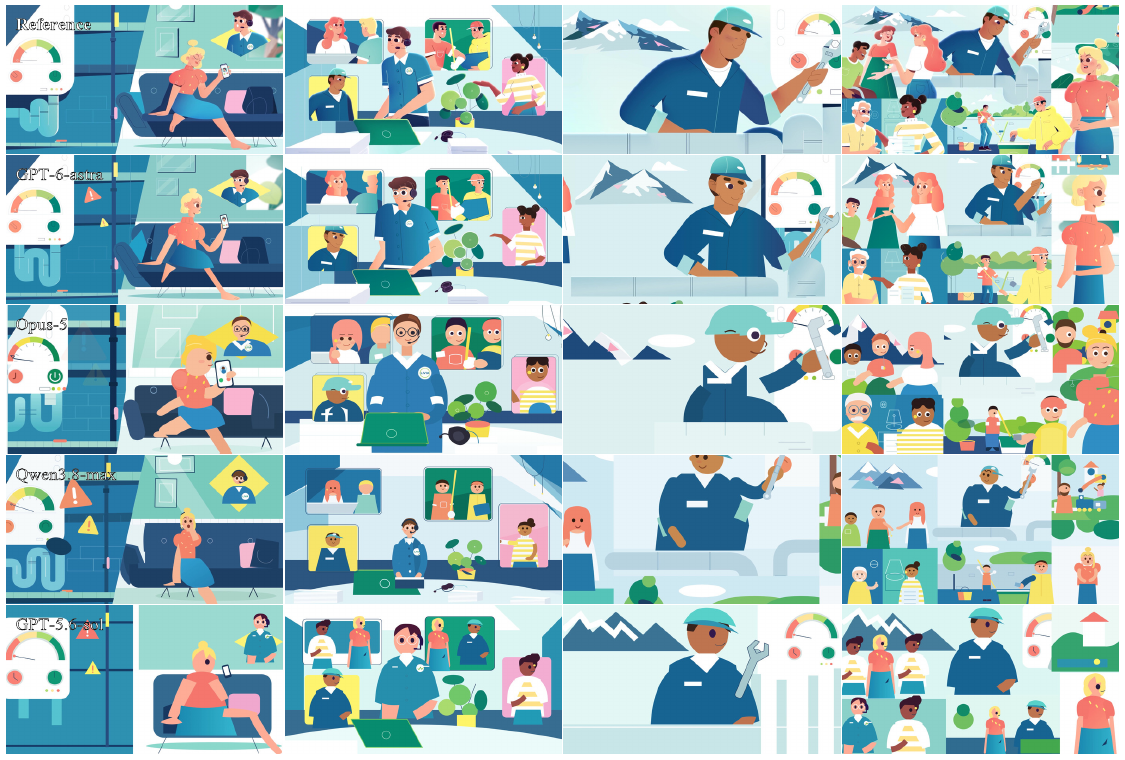}
    \caption{Character and scene animation. The four columns correspond to 7.7, 9.6, 13.3, and 15.6 seconds, showing character appearance, scene composition, and transitions.}
    \label{fig:appendix_case_c}
\end{figure*}

\begin{figure*}[htbp]
    \centering
    \includegraphics[width=\textwidth]{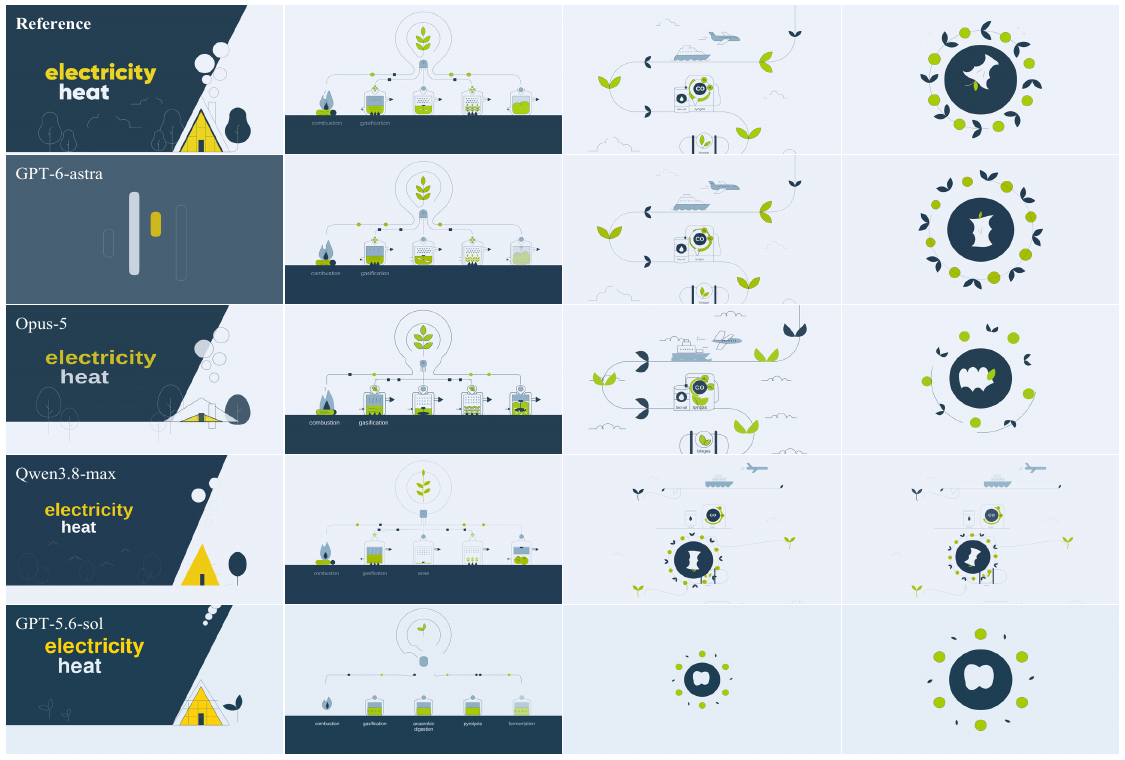}
    \caption{Scientific visualization. The four columns correspond to 0.5, 4.3, 26.5, and 27.3 seconds, showing energy conversion diagrams, process structure, and graphical transitions.}
    \label{fig:appendix_case_e}
\end{figure*}

\begin{figure*}[htbp]
    \centering
    \includegraphics[width=\textwidth]{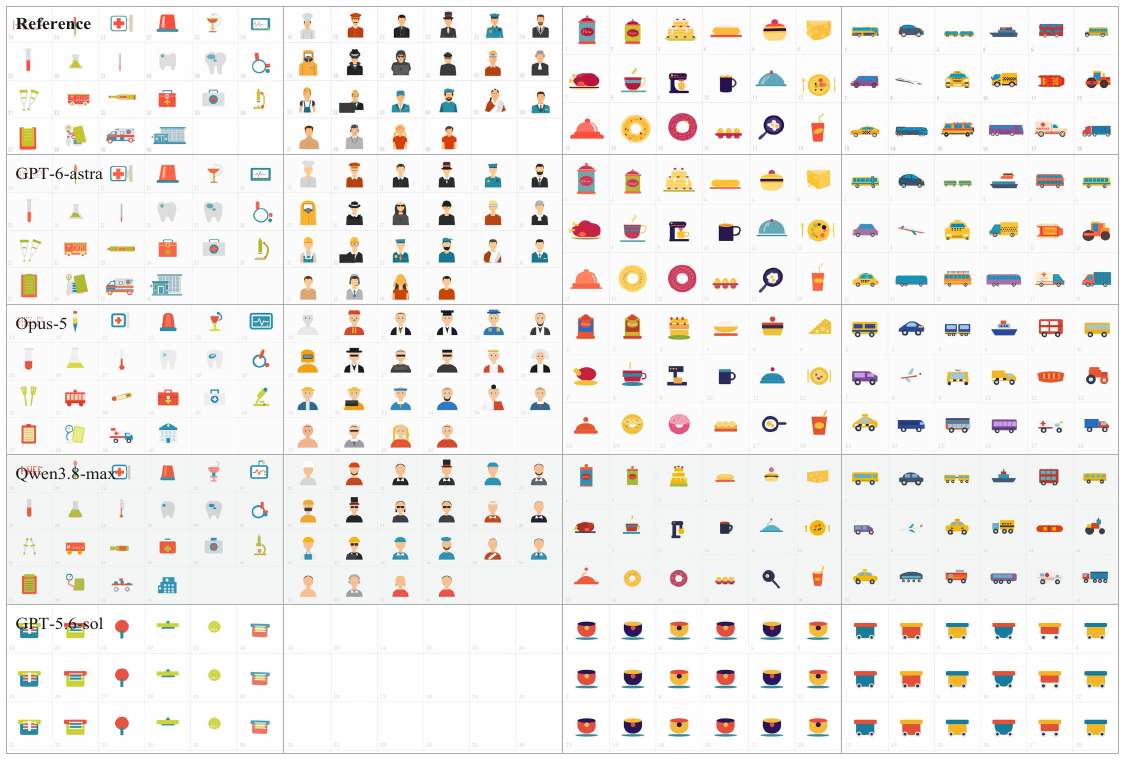}
    \caption{Icon collections. The four columns correspond to 3.0, 15.3, 17.5, and 23.0 seconds, showing the shapes, colors, and arrangement of different icon groups.}
    \label{fig:appendix_case_f}
\end{figure*}

\end{document}